\documentclass[10pt,twocolumn,letterpaper]{article}

\usepackage{3dvjb}
\usepackage{times}
\usepackage{epsfig}
\usepackage{graphicx}
\usepackage{amsmath}
\usepackage{amssymb}

\usepackage{color}
\usepackage{booktabs}
\usepackage{multirow}
\usepackage[colorlinks]{hyperref}
\usepackage{url}

\threedvfinalcopy 

\def\threedvPaperID{200} 

\begin{document}


\newif\ifdraft
\draftfalse

\definecolor{orange}{rgb}{1,0.5,0}

\ifdraft
 \newcommand{\PF}[1]{{\color{red}{\bf PF: #1}}}
 \newcommand{\pf}[1]{{\color{red} #1}}
 \newcommand{\JB}[1]{{\color{blue}{\bf JB: #1}}}
 \newcommand{\jb}[1]{{\color{blue} #1}}
 \newcommand{\MS}[1]{{\color{green}{\bf MS: #1}}}
 \newcommand{\ms}[1]{{\color{green} #1}}

\else
 \newcommand{\PF}[1]{}
 \newcommand{\pf}[1]{ #1 }
 \newcommand{\JB}[1]{}
 \newcommand{\jb}[1]{ #1 }
 \newcommand{\MS}[1]{}
 \newcommand{\ms}[1]{ #1 }
\fi

\newcommand{\comment}[1]{}
\renewcommand{\floatpagefraction}{.99}

\newcommand{\real}{\mathbb{R}}
\newcommand{\natur}{\mathbb{N}}
\newcommand{\x}{\mathbf{x}}
\newcommand{\y}{\mathbf{y}}
\renewcommand{\d}{\mathbf{d}}
\newcommand{\w}{\mathbf{w}}
\newcommand{\m}{\mathbf{m}}
\newcommand{\radius}{\mathbf{r}}
\newcommand{\C}{\mathbf{C}}

\newcommand{\Ib}{\mathbf{I}}
\newcommand{\Bb}{\mathbf{B}}
\newcommand{\Sb}{\mathbf{S}}
\newcommand{\Db}{\mathbf{D}}
\newcommand{\Nb}{\mathbf{N}}
\newcommand{\Mb}{\mathbf{M}}
\newcommand{\Kb}{\mathbf{K}}
\newcommand{\Pb}{\mathbf{P}}
\newcommand{\Ab}{\mathbf{A}}
\newcommand{\nb}{\mathbf{n}}
\newcommand{\Nbh}{\hat{\mathbf{N}}}
\newcommand{\vb}{\mathbf{v}}

\newcommand{\normeucl}[1]{\left\lVert #1 \right\rVert}

\newcommand{\OURS}[0]{\textbf{OURS}}

\title{Learning to Reconstruct Texture-less Deformable Surfaces from a Single View}

\author{Jan Bedna\v{r}\'ik \\
{\tt\small jan.bednarik@epfl.ch}
\and
Pascal Fua\\
{\tt\small pascal.fua@epfl.ch}
\and
Mathieu Salzmann\\
{\tt\small mathieu.salzmann@§epfl.ch}
}

\date{\'Ecole Polytechnique F\'ed\'erale de Lausanne, Switzerland}

\maketitle


\begin{abstract}
Recent years have seen the development of mature solutions for reconstructing deformable surfaces from a single image, provided that they are relatively well-textured. By contrast, recovering the 3D shape of texture-less surfaces remains an open problem, and essentially relates to Shape-from-Shading. In this paper, we introduce a data-driven approach to this problem. We introduce a general framework that can predict diverse 3D representations, such as meshes, normals, and depth maps. Our experiments show that meshes are ill-suited to handle texture-less 3D reconstruction in our context. Furthermore, we demonstrate that our approach generalizes well to unseen objects, and that it yields higher-quality reconstructions than a state-of-the-art SfS technique, particularly in terms of normal estimates. Our reconstructions accurately model the fine details of the surfaces, such as the creases of a T-Shirt worn by a person.
\end{abstract}


\section{Introduction} \label{sec:introduction}

\begin{figure*}[t]
	\centering
	\includegraphics[width=0.8\linewidth]{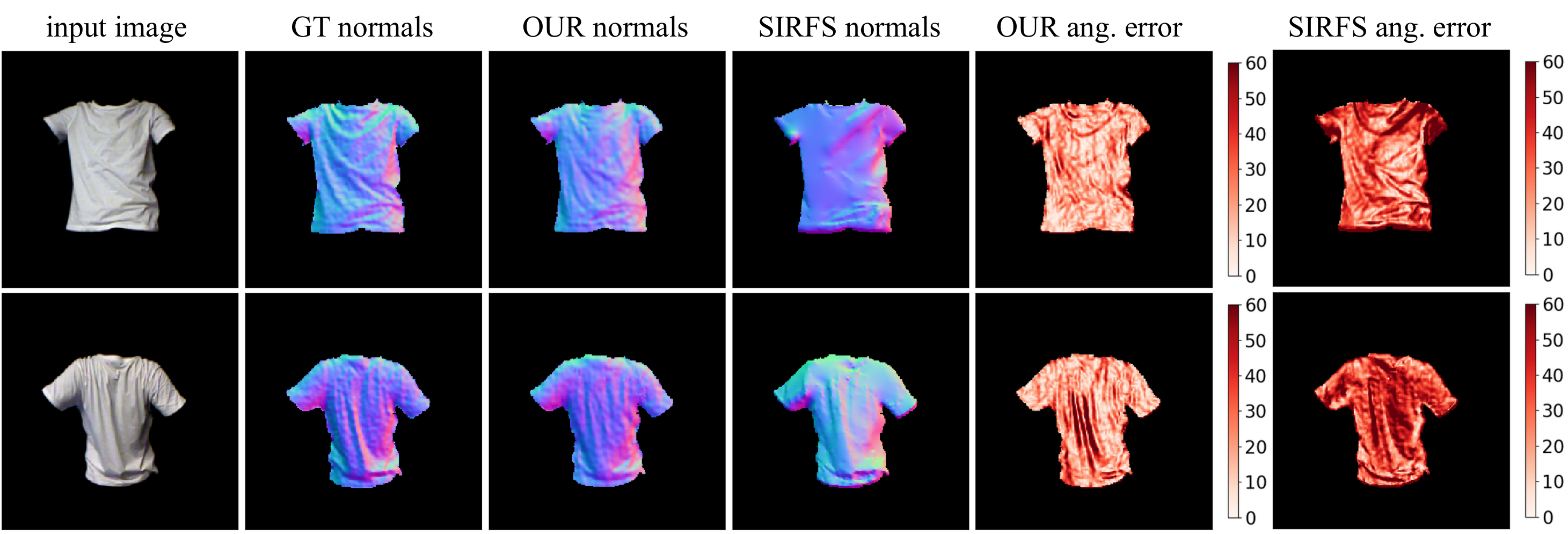}
	\caption{{\bf Reconstructing a T-Shirt.} Comparison of our method to that of~\cite{Barron15} on a deforming T-Shirt. The second to fourth columns depict the ground truth and recovered normals and the last two columns show the angular error in degrees. Note that the discrepancies are much smaller in our case and that the sharp creases are better recovered.}
	\label{fig:teaser}
	\vspace{-0.4cm}
\end{figure*}

In this paper, we tackle the problem of recovering the shape of complex and deforming texture-less surfaces from a single image, which is close in spirit to Shape-from-shading (SfS) with the added difficulty that we must handle complex phenomena such as sharp creases and self-shadowing. The T-shirt of Fig.~\ref{fig:teaser} being worn by someone who moves illustrates that.
This is in contrast to recent approaches that focus on well-textured surfaces~\cite{Ngo16,Bartoli12b}, or partially textured ones~\cite{Varol12b, Wang16e,Pumarola18}, and tend to produce coarse reconstructions in which fine details are lost.

SfS is one of the oldest Computer Vision problems~\cite{Horn89,Zhang99,Durou08}. Yet to this day, it remains largely unsolved because it is such  an ill-posed inverse problem, except in tightly controlled lighting environments~\cite{Prados05}. The early methods were variatonal ones that required very strong assumptions about the world, such as the presence of a single light source together with simple reflectance properties of the surfaces to be reconstructed, which are rarely satisfied. Recent ones~\cite{Barron15,Xiong15,Richter15} have focused on replacing some of these assumptions by measurements of the surface and lighting properties, yet still rely on relatively simple geometric and photometric models to remain computationally tractable. 

In this paper, we show that a data driven approach enables us to operate under much weaker assumptions that {\it are} sufficiently well satisfied in everyday life to make the method truly practical in an environment where the lighting can be complex and inter-reflections, shadows, and sharp creases are prevalent. Fig.~\ref{fig:teaser} depicts such a situation in which we outperform one of the best currently available algorithms~\cite{Barron15}.  It would seem natural to follow the most popular trend in modeling deformable surfaces and to train a Deep Net to regress from the image to the shape parameters of a surface mesh, as was done for well-textured surfaces in~\cite{Pumarola18}. We will demonstrate, however, that this is not the best approach. It is more effective to train a network that predicts a dense map of depths, normals, or both, as was done by the SfS pioneers~\cite{Horn89}. 

Because we do not constrain the surface to be smooth and allow the network to learn about complex effects such as self-shadowing and occlusions, we can recover very severe deformations such as the sharp folds that can be seen in Fig.~\ref{fig:teaser}.  Furthermore, as evidenced by our experiments, training on a single surface allows us to generalize to other ones of different shape and without any re-training or fine-tuning. Our two main contributions are: (i) A data-driven SfS approach that can recover much more complex deformations than earlier ones under realistic lighting conditions and which, unlike state-of-the-art intrinsic image decomposition techniques, only requires supervision in the form of depth and normal maps which are relatively easy to obtain. (ii) A large annotated real-world dataset consisting of $26500$ samples of surfaces with uniform reflectance undergoing complex deformations and viewed under complex realistic lighting. We thoroughly evaluate the performance of our method and show that it outperforms the state-of-the-art SfS approach of~\cite{Barron15}, whose code is available on the web.


\section{Related Work} \label{sec:related_work}

Traditionally, the SfS problem has been posed as a variational problem involving the optimization of physically-inspired objective functions to impose brightness, smoothness, and integrability constraints. In its original form~\cite{Horn89,Zhang99,Durou08}, the problem is underconstrained and its solution plagued by ambiguities~\cite{Belhumeur99,Ecker10}, which can be formally resolved only in very specific cases, such as when the camera and light source are co-located~\cite{Prados05} or when additional stereo information is available~\cite{Samaras00}. 

Known lighting, and absence of interreflections and cast shadows are often assumed, as in the approach of~\cite{Barron11} to jointly recover albedo and shape so as to explain the image as well as possible. In~\cite{Oxholm12,Johnson11}, while known, the lighting is assumed to be natural, which makes it possible to treat the thee color channels of the image in a manner similar to that of photometric stereo. By contrast, our model does not require any prior knowledge about the lighting.

Assumptions are also often made about the shape. For example, in~\cite{Xiong15}, quadratic functions are fitted to local image patches of different sizes, which allows the prediction of normals if the surface is sufficiently smooth, while in~\cite{Hassner06,Huang07}, exemplars are used to provide shape priors. A different approach is to assume the direction of the normals to be correlated with their distance to the occlusion boundary \cite{Richter15}, or to learn the smooth shape priors directly from data \cite{Barron15}. If the object category is known, sparse parameterization can be used instead of dense depth/normal maps or meshes. For instance, human face reconstruction approaches often rely on 3D Morphable Models~\cite{Richardson17,Shu17a,Jourabloo16,Bagautdinov18}.

Recently, the most popular strategy has been to jointly infer two or more modalities that contribute to the image formation process, specifically normal map, depth map, surface reflectance, reflectance map and/or lighting parameters in either optimization based \cite{Barron15,Oxholm12,Zoran14,Choe16} or learning based \cite{Richter15,Rematas16,Sengupta18,Janner17,Shu17a} setting. This has been one of our motivations for developing a multi-stream CNN model that outputs multiple shape representations, as will be discussed in Section~\ref{sec:method_overview}. Even though we were inspired by Deep Net based models performing intrinsic image decomposition, our method relaxes some of the rather restricting assumptions and need for hard-to-obtain annotations. Specifically, \cite{Richter15} assumes a Lambertian reflectance model under Spherical Harmonics lighting, which is rarely the case in practice. \cite{Sengupta18,Janner17} require GT albedo and lighting annotations while \cite{Shu17a} focuses solely on the human face object category as it relies on 3DMM representation. In \cite{Rematas16}, the surface normals are inferred as the by-product of reflectance map estimation, however, only low resolution of $64\times64$ px is supported and the results are only reported on synthetic data coming from a single object category. Another Deep Net based approach to directly predicting a normal map from an input image was introduced in~\cite{Yoon16} but it can only operate on infrared input images, whereas our approach takes a standard RGB image as input. 

In our approach, we  do not attempt to recover the lighting or reflectance explicitly, since such measurements are difficult to obtain ground-truth annotations for. Instead, we let the network learn how to handle these quantities from data. As evidenced by our experiments, even without explicitly modeling or predicting lighting and material BRDF, our network can successfully reconstruct fine surface details of complex shapes acquired under realistic conditions.

In the context of monocular 3D reconstruction of deformable surfaces, the most recent methods rely on CNNs to regress from the image to mesh vertices~\cite{Pumarola18,Danerek17}. 
However, while~\cite{Pumarola18} can recover complex deformations, it focuses on well-textured surfaces. By contrast,~\cite{Danerek17} handles poorly-textured surfaces, but visual inspection of their results clearly shows that the method oversmoothes the shape. Our approach focuses on texture-less objects, and, as depicted by our results in Fig.~\ref{fig:teaser}, is able to reconstruct fine-grained deformations, such as the creases of a T-Shirt. Furthermore, we show that mesh representations are outperformed by normal- and depth-based ones for this task.


\section{Our Approach} \label{sec:method_overview}

\begin{figure*}
	\centering
	\includegraphics[width=0.65\linewidth]{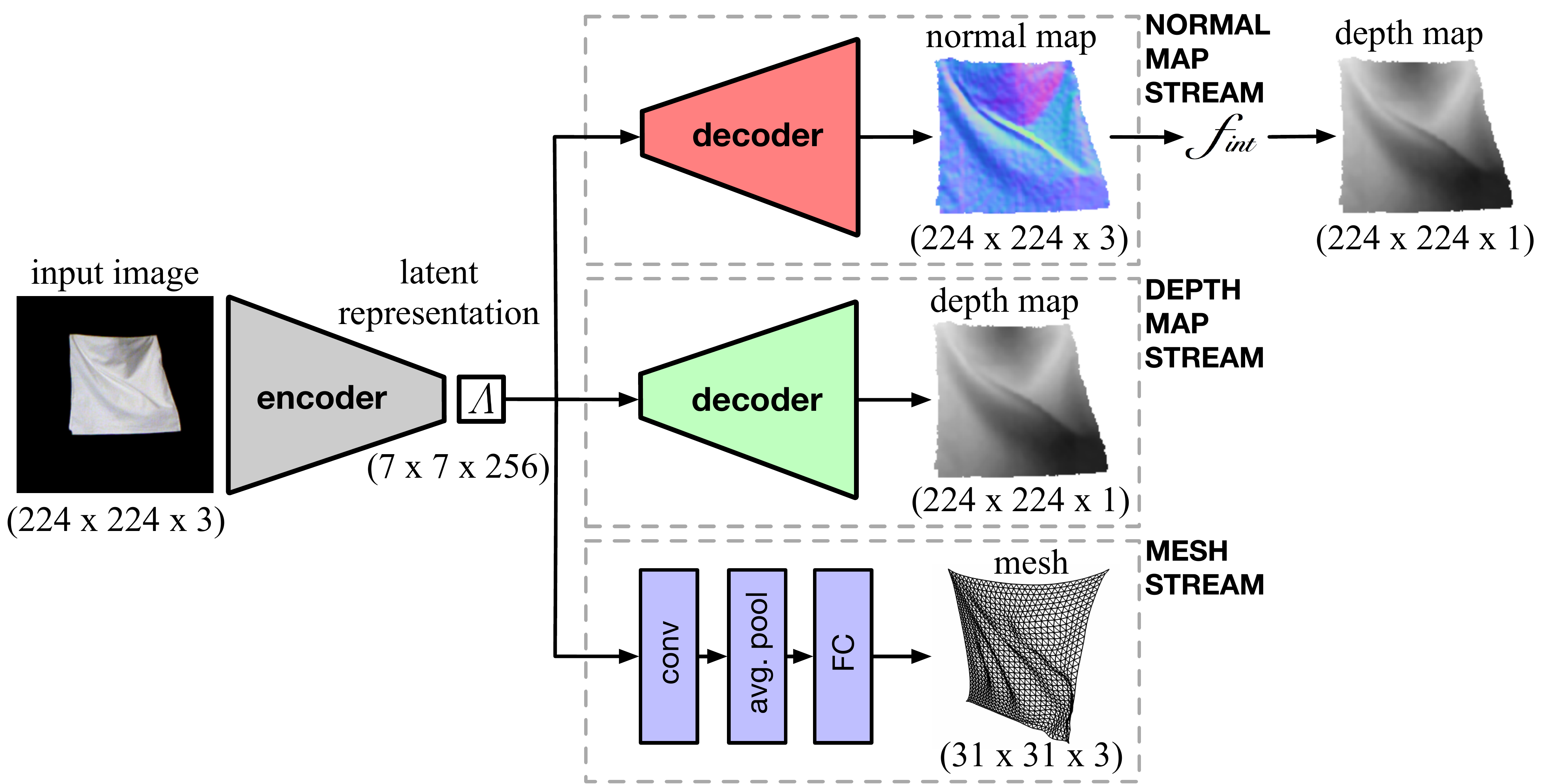}
	\caption{{\bf Surface reconstruction architecture.} Our model is based on SegNet~\cite{Badrinarayanan15}, but here we consider multiple output branches. The encoder, in gray, outputs the same latent representation $\Lambda$ for the normal, depth, and vertex branch. A different decoder, shown in red, green, and blue, is then used for each branch. In other words, this creates three potential streams, which can be either trained individually or jointly. When recovering normals only, an additional integration step is required to compute depths.}
	\label{fig:architecture}
	\vspace*{-0.3cm}
\end{figure*}

\subsection{Problem Formulation} 
\label{sec:formulation}

Let $\Ib \in \real^{H \times W \times 3}$ be an RGB image of size $W \times H$ and $\Bb \in \natur_{0}^{H \times W \times 3}$ a binary mask that denotes the foreground region to be recovered. Our goal is to learn a mapping $f_{SR}: \Ib \odot \Bb \rightarrow \Sb$, where $\Sb$ represents the corresponding 3D surface.  For deformable surfaces, a natural 3D representation would be a vector $\Sb_M$ containing the 3D vertex coordinates of a triangulated mesh, as in~\cite{Pumarola18}. However, other representations such as a depth map $\Sb_D$ or a normal map $\Sb_N$, which are more prevalent in SfS papers, can also be used. In fact, these representations are not mutually exclusive, and we can train a network to return one or more of them, as shown in Fig.~\ref{fig:architecture}. In this section, we discuss this general scenario. However, in practice, we have found that the mesh-based representation was not as effective as the other two. 

Given a calibrated camera, a triangulated mesh with V vertices can be expressed as a vector $\Sb_M \in \real^{3V}$ of 3D points in the camera coordinate frame. By contrast, a depth map and a normal map can be encoded as images instead of vectors. Specifically, the depth map $\Sb_D$ is a $W \times H$ floating point image, and the normal map $\Sb_N$ is a three-channel floating point image of size $W \times H \times 3$  representing the $x$, $y$ and $z$ coordinates of the normal vector expressed in the camera coordinate frame. Training depth maps can be easily acquired using existing depth sensors, such as  the Microsoft Kinect camera. This data can be converted into ground-truth normal maps by smoothing and differentiating the depth maps as discussed in Section~\ref{sec:data_processing}. By contrast, obtaining training data for 3D meshes for real images is harder and requires much more processing, since depth sensors do not provide correspondences between points on the surface. However, unlike the other two representations, 3D meshes can represent self-occluded parts of the surface, albeit at the cost of constraining the topology much more. We explain the process of obtaining the GT mesh coordinates in the supplementary material.
 
\subsection{Shape Recovery Networks} 
\label{sec:network}

In this work, we rely on the SegNet deep autoencoding architecture  \cite{Badrinarayanan15} depicted by Fig.~\ref{fig:architecture} to regress from the image to each of our three representations.

Let $\Ib_{m}=\Ib \odot \Bb$ be the foreground image. The encoder performs feature extraction and outputs a latent representation tensor $\Lambda(\Ib_{m}) \in \real^{H_{L} \times W_{L} \times C_{L}}$, whose spatial size is $(H_{L} \times W_{L})$ and third dimension $C_{L}$ is the number of filters in the last convolutional layer of the encoder. As in many state-of-the-art intrinsic image decomposition approaches~\cite{Janner17,Shi06,Shu17a}, we assume the learned features are independent of the final output modality, and thus we use the same encoder for all three shape representations. Keeping the encoder design and shape of latent representation $\Lambda(\Ib_{m})$ the same for all three scenarios allows us to not only train each model separately but also jointly, which helps the model learning more robust feature extractors, which results in higher reconstruction accuracy as shown in Section~\ref{sec:results}. Note, however, that the weights of this encoder will differ if we train it, for instance, for depth map prediction only or for mesh prediction only.

Let $\Psi_{C}$, $\Psi_{D}$, $\Psi_{N}$ be the decoders for mesh vertices, depth, and normals, respectively. Inspired by branched Deep Net architectures, which have been shown to perform well for intrinsic image decomposition~\cite{Janner17,Shu17a} and multi-task learning~\cite{Li15d}, we do not force the design of the decoders to match each other but rather adjust them to suit the output shape and/or topology. 

Since the depth and normal maps both have an image-like topology, as does the output of the original SegNet, we use the same SegNet decoder architecture for these two modalities, except for the number of convolutional filters in the last convolutional layer, that is, 1 for depth and 3 for normals, and for the fact that these outputs do not need to be passed through a softmax. By contrast, when using the mesh representation, the output size is significantly smaller and shaped as a vector. We therefore take $\Psi_{C}$ to be a single convolutional layer followed by average pooling and a fully connected layer to regress to the vertices' coordinates.

\subsection{Loss Functions} 
\label{txt:loss_functions}

Let us assume to be given $N$ training samples. We represent each one as a tuple $(\Ib^n,\Bb^n,\vb^n,\Db^n,\Nb^n)$, that is,  an input image $\Ib^n$ with corresponding foreground mask $\Bb^n$, ground-truth mesh vertices $\vb^n$, depth map $\Db^n$ and normal map $\Nb^n$. We define 3 loss functions for the three potential outputs of our network.

To train $\Psi_{C}$, we define the loss as the Mean Square Error between the vertex coordinates and the ground truth. That is,
\begin{align} \label{eq:vertex-wise-MSE}
\mathcal{L}_{C} = \frac{1}{N} \sum_{n=1}^N\frac{1}{V} \sum_{i = 1}^{V}\normeucl{\vb_{i}^{n} - \Psi_{C}(\Lambda{(\Ib_{m}^n)})_{i}}^2\;,
\end{align}
where a subscript $i$ denotes the vertex number.

Since we use training data whose average distance to the camera is roughly constant and focus on recovering local high-frequency deformations, to train $\Psi_{D}$,  we minimize the loss
\begin{align} \label{eq:pixel-wise-L1}
\mathcal{L}_{D} = \frac{1}{N}\sum_{n=1}^N\frac{\sum_{i}{|\Db_{i}^{n} - \Psi_{D}(\Lambda(\Ib_{m}^n))_{i}|\Bb_{i}^n}}{\sum_{i}{\Bb_{i}^n}} \; ,
\end{align}
where $i$ denotes the image location. Note that we only take into account pixels within the binary mask $\Bb$. In other words, we handle the depth ambiguity by recovering depth variations around the mean depth of our training data, instead of using a scale invariant measure as in~\cite{Eigen14}. As will be discussed in Section~\ref{sec:results}, to evaluate accuracy at test time, we rescale the prediction so as to align it with the ground truth.

Similarly, to train $\Psi_{N}$, we define a loss $\mathcal{L}_{N}$ that relies on a linearized version of the cosine similarity~\cite{Trigeorgis17} and add to it a term that favors unit length vectors. We take it to be
\begin{align}
\mathcal{L}_{N} = \frac{1}{N}\sum_{n=1}^N \frac{\left[ \sum_{i}\Bb_i^n \left(\kappa \mathcal{L}_{a}(\Nb_{i}^{n}, \Nbh_{i}^{n}) + \mathcal{L}_{l}(\Nbh_{i}^{n})\right) \right]}{\sum_{i}{\Bb_{i}^n}}\;,
\label{eq:loss_ange}
\end{align}
with
\begin{align}
\mathcal{L}_{a}(\Nb_{i}^{n}, \Nbh_{i}^{n}) &= \arccos \left( \frac{ \Nb_{i}^{n} \Nbh_{i}^{n} }{\|\Nb_{i}^{n}\|\|\Nbh_{i}^{n}\| + \epsilon} \right) \frac{1}{\pi}\;, \\
\mathcal{L}_{l}(\Nbh_{i}^{n}) &= \left(\normeucl{\Nbh_{i}^{n}} - 1\right)^{2},
\end{align}

where $\Nbh^{n} = \Psi_{N}(\Lambda(\Ib_{m}^n))$ denotes the predicted normal map, $\epsilon$ is a small positive constant that prevents divisions by zero and increases numerical stability, and $\kappa$ sets the relative influence of the two terms in the loss function. In our experiments, we chose $\kappa=10$.


\newcommand{\cloth}[0]{\texttt{cloth}}
\newcommand{\tshirt}[0]{\texttt{tshirt}}
\newcommand{\sweater}[0]{\texttt{sweater}}
\newcommand{\hoody}[0]{\texttt{hoody}}
\newcommand{\paper}[0]{\texttt{paper}}

\section{Real-World SfS Dataset} \label{sec:real-worl_dataset_acquisition}

Successful training of most of the Deep Net models depends on access to large training databases. By contrast, the datasets used in the SfS literature remain relatively small. For example, the algorithm of~\cite{Xiong15} relies on 7 rigid objects under 20 different directional lighting. In~\cite{Barron12a}, an augmented version of the MIT intrinsic image dataset~\cite{Grosse09} containing 20 rigid objects under various illumination is used while the method of~\cite{Khan09b} works with a database of 6 human faces. Some authors use only synthetic data~\cite{Richter15} while others augment the limited amount of real-data with synthetic data~\cite{Xiong15,Barron15}. 

By contrast and to fully exploit the capacity of our model while avoiding overfitting, we captured a new large dataset of real deforming surfaces. We acquired sequences of RGB images and corresponding depth maps of a rectangular piece of cloth (\cloth{}), T-shirt (\tshirt{}), sweater (\sweater), hoody (\hoody{}) and crumpled sheet of paper (\paper{}) undergoing complex deformations and seen under varying lighting conditions. We chose the cloth for two reasons. First, being a generic piece of cotton fabric makes it universal enough to capture a wide distribution of local deformations and appearance, even if it differs globally from other objects such as garments. Second, its flat rest state makes it easy to represent by a triangular mesh, as is often done in deformable surface reconstruction~\cite{Salzmann11a,Bartoli12b,Ngo16,Pumarola18}, and to test our mesh-based model. By contrast, the pieces of garment and a sheet of paper were chosen as a more complex real-world object to demonstrate the generality of our approach and the fact that training on \cloth{} produces good results on the \tshirt{}, \sweater{}, \hoody{} and \paper{}.

The resulting dataset comprises a total of $26500$ image-normals-depth triplets, which is much larger than existing real-world SfS datasets, and we make it publicly available\footnote{\url{https://cvlab.epfl.ch/texless-defsurf-data}}. We now describe the acquisition process.

\subsection{Acquisition Setup}

\begin{figure}[t]
	\centering
	\includegraphics[width=1.0\linewidth]{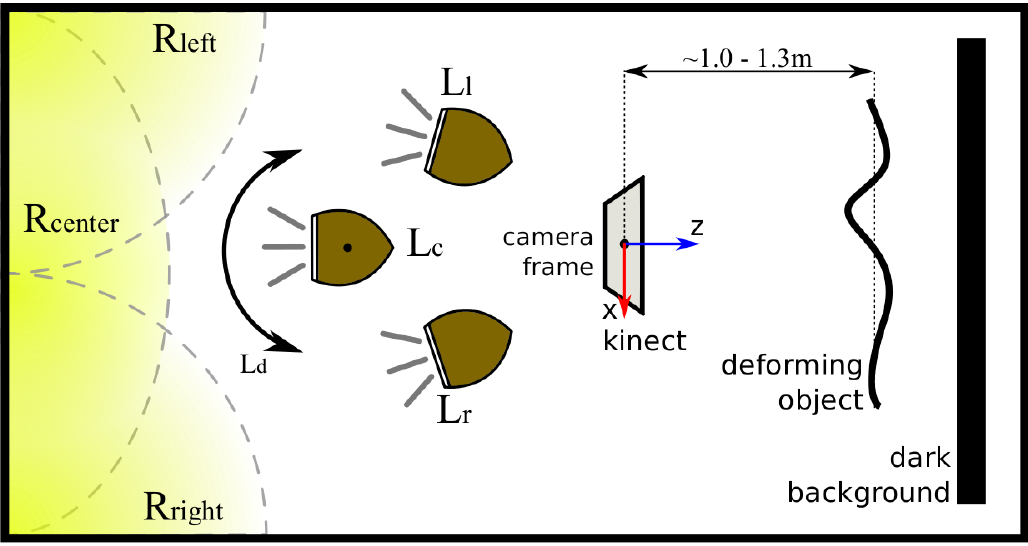}
	\caption{{\bf Data acquisition setup.} The deformable surface is placed between a Microsoft Kinect camera and a dark background. We use three static light sources, $L_{r}$, $L_{l}$ and $L_{c}$, pointing towards the back wall and a fourth one, $L_{d}$, which can move. The deformable surface is therefore lit by complex indirect lighting.}
	\label{fig:room}
	\vspace*{-0.3cm}
\end{figure}

Fig.~\ref{fig:room} illustrates our setup. We placed a given deformable object in front of a dark background and captured synchronized RGB images and depth maps using a Microsoft Kinect camera, positioned so that its optical axis is roughly perpendicular to the background plane.

\begin{figure}[t]
	\centering
	\includegraphics[width=0.9\linewidth]{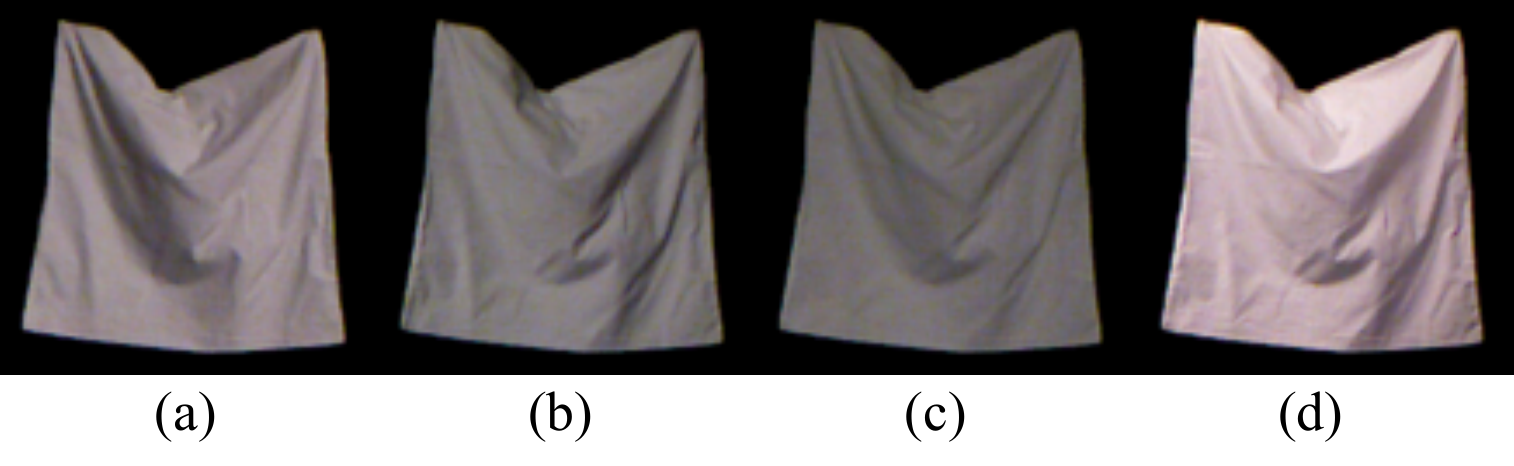}
	\vspace{-0.2cm}
	\caption{{\bf Shading effects on a rectangular piece of cloth.} We show the effects of (a) $L_{r}$, (b) $L_{l}$, (c) $L_{c}$ and (d) a randomly chosen frame for $L_{d}$.}
	\label{fig:lightings}
	\vspace*{-0.5cm}
\end{figure}

We used three fixed incandescent lamps positioned in the right, left and central area of the room, and a fourth one, that can move. All four were slightly slanted upwards and pointed towards the back of the room, that is, in the direction opposite to the camera's optical axis. As a result, the deforming surface was mostly illuminated by light reflecting off the walls and coming from the {\it radiance regions} shown in yellow in Fig.~\ref{fig:room}. This setup simulates directional but diffuse and soft lighting. The range of motion of the dynamically moving light source was chosen so that its radiance regions were slightly larger than those of the other three put together, which should result in a richer light distribution, as shown in  Fig.~\ref{fig:lightings}. In the remainder of this section, we will consider 4 separate scenarios in which we use each lamp individually and will refer to them as $L_{r}$, $L_{l}$, $L_{c}$, and $L_{d}$, where the subscripts $r, l, c$ and $d$ stand for \textit{right}, \textit{left}, \textit{central} and \textit{dynamic},  respectively.

The Microsoft Kinect camera was calibrated, thus yielding known camera intrinsics. Depth maps and corresponding RGB images were aligned using OpenKinect libfreenect library. We recorded sequences at 5 FPS to avoid capturing too many duplicates of nearly identical shapes. Each recorded frame consists of a 640 x 480 RGB image and a 640 x 480 depth map. Because of the deformation undergone by the surfaces, some of the frames feature substantial motion blur, which further increases both realism and the challenge.

\subsection{Deforming Surfaces}
\vspace{-0.1cm}
To acquire the images, we pinned the rectangular cloth to a fixed bar along a given edge or corners and manually deformed the rest. In contrast, the T-shirt, the sweater and the hoody were worn by different people making random body motions. The T-Shirt was captured separately from the front and from the back. We were manually deforming the crumpled sheet of paper.

\begin{figure}[t]
	\centering
	\includegraphics[width=1.0\linewidth]{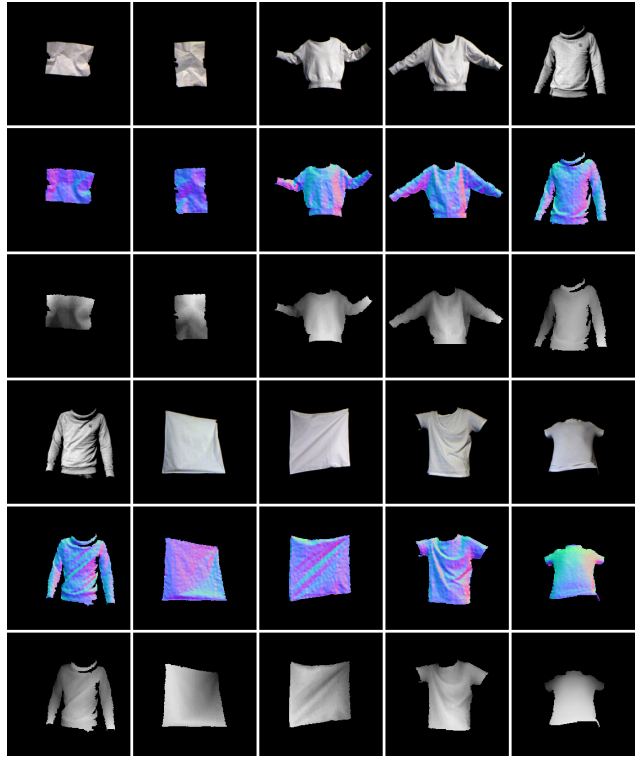}
	\caption{{\bf Randomly chosen samples from our dataset.} RGB images together with corresponding GT normals and depth maps are shown for \paper{}, \sweater{}, \hoody{}, \texttt{cloth} and \texttt{tshirt}.}
	\label{fig:ds}
	\vspace*{-0.4cm}
\end{figure}

Altogether, this resulted in $18$ sequences of $15799$ samples for \cloth{}, $12$ sequences of $6739$ samples for \tshirt{}, $4$ sequences of $2203$ samples for \sweater{}, $1$ sequence of $517$ samples for \hoody{} and $3$ sequences of $1187$ samples for \paper{}. A representative set of samples from our final dataset is depicted in Fig.~\ref{fig:ds}

\subsection{Data Preprocessing} \label{sec:data_processing}

As explained in Section~\ref{sec:method_overview}, we assume the foreground binary mask $\Bb$ to be known for each image. To create it, we segmented the imaged objects in the RGB images by simple thresholding followed by hole filling of the biggest connected component. The masks were then used to segment the corresponding depth maps. Furthermore, the RGB images were white balanced using a standard color checker to account for the often unsatisfactory automatic white balance of the Kinect camera. Both the RGB images and depth maps were cropped to the spatial size of 224 x 224, which is the expected input/output size of our model. Since the raw depth maps contain noise and holes, we performed distance based clustering and hole filling by interpolation. The normal maps were computed by differentiating the depth maps in a finite difference sense. Since the Kinect depth measurements are noisy, before computing the normal maps, we smoothed the depth maps by applying a Gaussian blur kernel of size $9\times 9$ with $\mu = 0, \sigma=3$. As is apparent in Fig.~\ref{fig:ds}, there is still some noise in the normal maps. However, its magnitude is low enough not to significantly corrupt the fine-detailed high-frequency surface geometry, that is, the wrinkles that we are trying to model.


\newcommand{\norms}[0]{\textbf{Normals}}
\newcommand{\depth}[0]{\textbf{Depth}}
\newcommand{\coord}[0]{\textbf{Coords}}
\newcommand{\jointND}[0]{\textbf{N+D}}
\newcommand{\jointNC}[0]{\textbf{N+C}}
\newcommand{\jointDC}[0]{\textbf{D+C}}
\newcommand{\jointNDC}[0]{\textbf{N+D+C}}
\newcommand{\barron}[0]{{\bf Barron}}
\newcommand{\mae}[0]{$m_{AE}$}
\newcommand{\dae}[0]{$d_{AE}$}
\newcommand{\mc}[0]{$m_{C}$}
\newcommand{\md}[0]{$m_{D}$}

\section{Experiments and Results}
\label{sec:results}

Recall from Section~\ref{sec:network} that we can use the architecture depicted by~Fig.~\ref{fig:architecture} to recover normals, depths, or vertex coordinates either independently or jointly. In this section, we focus on independent recovery of the three modalities, which we will refer to as \norms{}, \depth{}, and \coord{}, respectively, joint recovery of two modalities, which we will denote \jointNC{}, \jointDC{} and \jointND{}, respectively, and joint recovery of all three modalities denoted as \jointNDC{}, where letters \textbf{N}, \textbf{D} and \textbf{C} denote the use of the normals decoder $\Psi_{N}$, depth decoder $\Psi_{D}$ and/or mesh decoder $\Psi_{C}$ in the final model. Note that for models predicting multiple modalities, we can evaluate the error for each individual output. We denote this by, e.g., \jointNDC{}/$X$ where $X$ 
can be $\Nb$, $\Db$, or \textbf{C}, depending on whether we evaluate with respect to the normals, depths, or coordinates.

We train our networks using the \cloth{} and \tshirt{} datasets under the four lighting scenarios introduced in Section~\ref{sec:real-worl_dataset_acquisition}. Each dataset includes separate training and testing sequences, and we can either train and test on the same object or train on one and test on the other. We can also train with one particular set of lights and test with a different one. In both cases, this allows us to gauge the generalization abilities of our approach. 

\begin{table}[t]
	\centering
	\resizebox{0.48\textwidth}{!}{
	\begin{tabular}{lcccc}
		\textbf{Experiment} & \textbf{Train obj.} & \textbf{Train light.} & \textbf{Test obj.} & \textbf{Test light.} \\
		\midrule
		cloth-cloth & cloth & $L_{r}$, $L_{l}$, $L_{d}$ & cloth & $L_{c}$ \\
		tshirt-tshirt & tshirt & $L_{r}$, $L_{c}$, $L_{d}$ & tshirt & $L_{l}$ \\
		cloth-tshirt & cloth & $L_{r}$, $L_{l}$, $L_{c}$, $L_{d}$ & tshirt & $L_{r}$, $L_{l}$, $L_{c}$, $L_{d}$ \\
		cloth-sweater & cloth & $L_{r}$, $L_{l}$, $L_{c}$, $L_{d}$ & sweater & $L_{r}$, $L_{l}$, $L_{c}$, $L_{d}$ \\
		cloth-hoody & cloth & $L_{r}$, $L_{l}$, $L_{c}$, $L_{d}$ & hoody & $L_{l}$ \\
		cloth-paper & cloth & $L_{r}$, $L_{l}$, $L_{c}$, $L_{d}$ & paper & $L_{r}$, $L_{l}$, $L_{c}$ \\
	\end{tabular}
	}
	\vspace*{0.0cm}
	\caption{{\bf List of the experiments we conducted.}
	}
	\label{tab:experiments_description}
	\vspace*{-0.5cm}
\end{table}%

Table~\ref{tab:experiments_description} summarizes the experiments we have conducted and whose results we report below. For each one, we randomly select $100$ samples from the test sequences and report results on. 

\subsection{Implementation Details} \label{txt:implementation_details}

For all the experiments described in this section, we used the Adam \cite{Kingma15} optimizer to train the network. We use a fixed learning rate of $0.001$ and parameter $\kappa$ of the loss function of Eq.~\ref{eq:loss_ange} set to $10$. For \norms{}, \depth{} and \coord, we simply let the optimization proceed. For \jointNC{}, \jointDC{} and \jointND{}, we started by training the model employing only one of the two decoders and we applied an early stopping that halted the training once the validation loss stopped decreasing for $30$ consecutive epochs. We then began estimating both decoders' outputs by minimizing the loss function $\mathcal{L}_{\text{joint}2} = \alpha \mathcal{L}_{d1} + \beta \mathcal{L}_{d2}$, where $\mathcal{L}_{d1}$ and $\mathcal{L}_{d2}$ each represent an appropriate loss function as defined in Eqs.~\ref{eq:vertex-wise-MSE}, \ref{eq:pixel-wise-L1} and \ref{eq:loss_ange} and where we fixed the mixing coefficients $\alpha = 1, \beta = 3$  to promote the training of the yet untrained decoder. As in the single-decoder scenario, we used the early-stopping technique. For \jointNDC, we similarly added the vertex-wise loss of Eq.~\ref{eq:vertex-wise-MSE} and continued training by minimizing the loss function $\mathcal{L}_{\text{joint}3} = \alpha \mathcal{L_{N}} + \beta \mathcal{L_{D}} + \gamma \mathcal{L_{C}}$ with mixing coefficients fixed to $\alpha = 1, \beta = 1, \gamma = 3$ to promote the training of the yet untrained mesh decoder. Our implementation relies on Keras with a Tensorflow backend.

\subsection{Metrics} \label{txt:metrics}

To evaluate the accuracy of the predicted mesh coordinates, we use the mean vertex-wise Euclidean distance, similar to the MSE of Eq.~\ref{eq:vertex-wise-MSE} we used to formulate the training loss, but without squaring the distances. We will refer to this metric as \mc{}.

Since the depth maps we produce are subject to an inherent global scale ambiguity~\cite{Eigen14},  we first align the corresponding point cloud to the ground truth using a Procrustes transformation~\cite{Stegmann02}. More precisely, let $\Theta_{\Kb}(\Ab)$ be the 3D point cloud associated to depth map $\Ab$ with corresponding intrinsic matrix $\Kb$, and let $\Omega(\Pb,\Db_b)$ denote the Procrustes transformation of cloud $\Pb$ with respect to depth map $\Db_b$. Given a set of $N$ ground-truth depth maps $\Db^n$ and predicted ones $\Delta^n$, we compute accuracy in terms of the metric
\begin{align}
m_{D} = \frac{1}{N}\sum_{n=1}^N\frac{\sum_{i}{\normeucl{\Theta_{\Kb}(\Db^{n})_{i} - \Omega(\Theta_{\Kb}(\Delta^{n}), \Db^{n})_{i}}\Bb_{i}^n}}{\sum_{i}{\Bb_{i}^n}},
\end{align}
where subscript $i$ indexes the pixels.
 
Finally, for normal maps, we  integrate the normals to recover the corresponding depth and again use the \md{} metric. We also report the mean and median angular errors, which we denote as \mae{} and \dae{}, respectively, as well as the fractions of normals exhibiting smaller angular error than $10^{\circ}$, $20^{\circ}$ and $30^{\circ}$. 

\subsection{Effectiveness of Meshes or the Lack Thereof}
\begin{figure}
	\centering
	\includegraphics[width=1.0\linewidth]{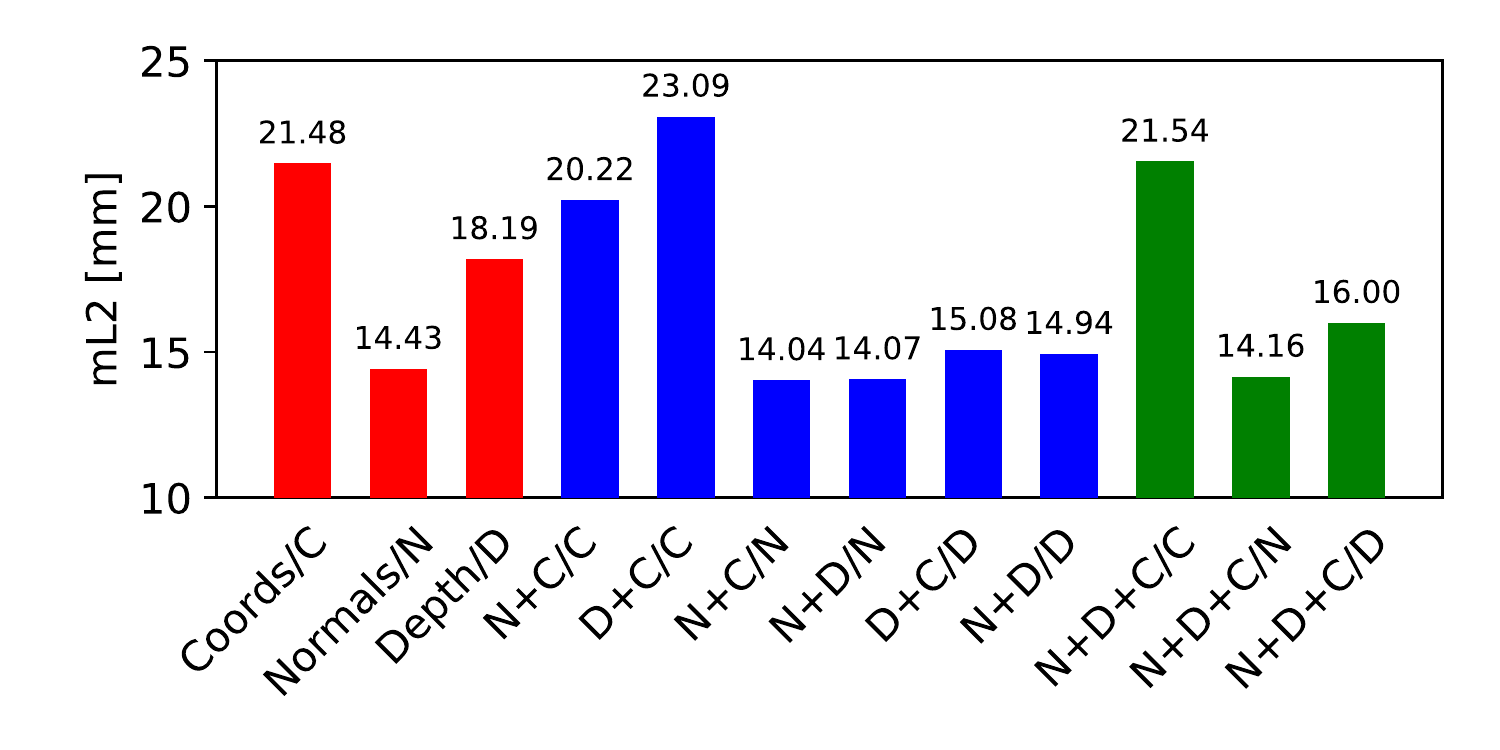}
	\caption{Comparison of models corresponding to all possible combinations of the normal map, depth map and mesh vertices decoders --- (red) single decoder models, (blue) two decoders models, (green) three decoders model. The models are trained and tested on \texttt{cloth} dataset and all their available inputs are considered for comparison using \mc{} and \md{} metrics.}
	\label{fig:modalities}
	\vspace*{-0.4cm}
\end{figure}

As mentioned in Section~\ref{sec:introduction}, our initial intuition was to follow the trend in deformable surface reconstruction and represent the surface as a triangular mesh. Here, we evaluate the results of such a representation, compared to depth and normal maps, and show that it is not as effective in our context. This is evidenced by the results in Figure~\ref{fig:modalities}, where we compare the models \coord{}, \norms{}, \depth{}, \jointNC{}, \jointDC{}, \jointND{} and \jointNDC{}, which we trained and tested on the \texttt{cloth} dataset and evaluated on all outputs available for the given model using metrics \mc{} for mesh coordinates and \md{} for depth maps. The predicted normal maps were converted to depth maps, i.e., we did not use normal maps themselves for this comparison.

We made three key observations: (1) In case of single modality models, i.e., \coord{}, \norms{} and \depth{}, the predicted mesh coordinates \coord{}/$\C$ yield by a large margin the highest error of $21.48$mm. (2) Training models \norms{} and \depth{} further using an additional modality, i.e., \jointNC{}, \jointDC{} or \jointND{}, in general helps reducing the error when producing depth or normal maps on the output. However, this is not the case for multi-decoder models producing mesh vertices, where the error can even further increase. (3) We found the model \jointNC{}/$\Nb$ to achieve overall the lowest error of $14.04$mm. However, this is comparable with \jointND{}/$\Nb$ and \jointND{}/$\Db$, which are not dependent on mesh vertices at all.

Given the non-trivial process required to create the GT mesh vertices, as described in the supplementary material, which is more tedious and error-prone than obtaining the depth or normal maps, and considering the negligible performance improvement, we discard the meshes from our approach in the remainder of the paper.

\subsection{Separate vs Joint Learning} \label{txt:separate_and_joint-streams_models}

In Table~\ref{tab:streams_comp}, we report the accuracy of \norms, \depth{}, and \jointND. In the latter case, we can evaluate either the predicted normals or the predicted depth maps, which we denote as \jointND/$\Nb$ and \jointND/$\Db$, respectively. To evaluate different scenarios we select \cloth{} and \tshirt{} from our datasets as the categories containing the most samples. We either train and test on the same object (\texttt{cloth}), or train on \texttt{cloth} and test on \texttt{tshirt}.

As can be seen in Table \ref{tab:streams_comp}, using the normal predictions, followed by integration, tends to yield lower errors than the predicted depth maps. More importantly, training jointly on normals and depth performs best overall, which is in keeping with the idea that forcing the network to learn features that disentangle the different contributions helps~\cite{Shu17a,Kulkarni15}. Interestingly, training on \cloth{} and testing on \tshirt{} degrades the accuracy but still yields a competitive result as we will see in the following section. 

\begin{table}[t]
	\centering
	\resizebox{0.47\textwidth}{!}{
		\begin{tabular}{lcc|cc}
			\textbf{Experiment} & \textbf{\jointND/$\Nb$} & \textbf{\norms/$\Nb$} & \textbf{\jointND/$\Db$} & \textbf{\depth{}/$\Db$} \\
			\midrule
			cloth-cloth & \textbf{17.53} & 17.80 & \textbf{15.96} & 18.18 \\
			tshirt-tshirt & 16.26 & \textbf{15.19} & \textbf{16.45} & 18.01  \\
			cloth-tshirt & \textbf{26.26} & 27.06 & \textbf{30.23} & 32.16 \\
		\end{tabular}
	}
	\caption{Comparison of \depth{}, \norms{} and \jointND{} in different scenarios. In general, the normal predictions yield lower \md{} error than the predicted depth maps, and joint training outperforms the single-decoder models.}
	\label{tab:streams_comp}
	\vspace*{-0.5cm}
\end{table}

\vspace*{-0.1cm}
\subsection{Comparing against a State-of-the-Art SfS Approach} \label{txt:comparison_with_state-of-the-art}

Here, we compare our results to those of the SIRFS method~\cite{Barron15}, which we briefly described in Section~\ref{sec:related_work}. This choice was motivated by the fact that SIRFS constitutes a state-of-the-art SfS method whose code is publicly available, unlike that of the other contemporary SfS methods described in that section. Given the input image and segmentation mask, SIRFS performs intrinsic image decomposition into a normal map, depth map, lighting and reflectance. To compare with our method we take SIRFS's normal map and depth map predictions, integrate the normals and align to the ground-truth depth map, as explained in Section \ref{txt:metrics} and we do the same for our own results. 

\begin{figure*}[!h]
	\centering
	\includegraphics[width=0.8\linewidth]{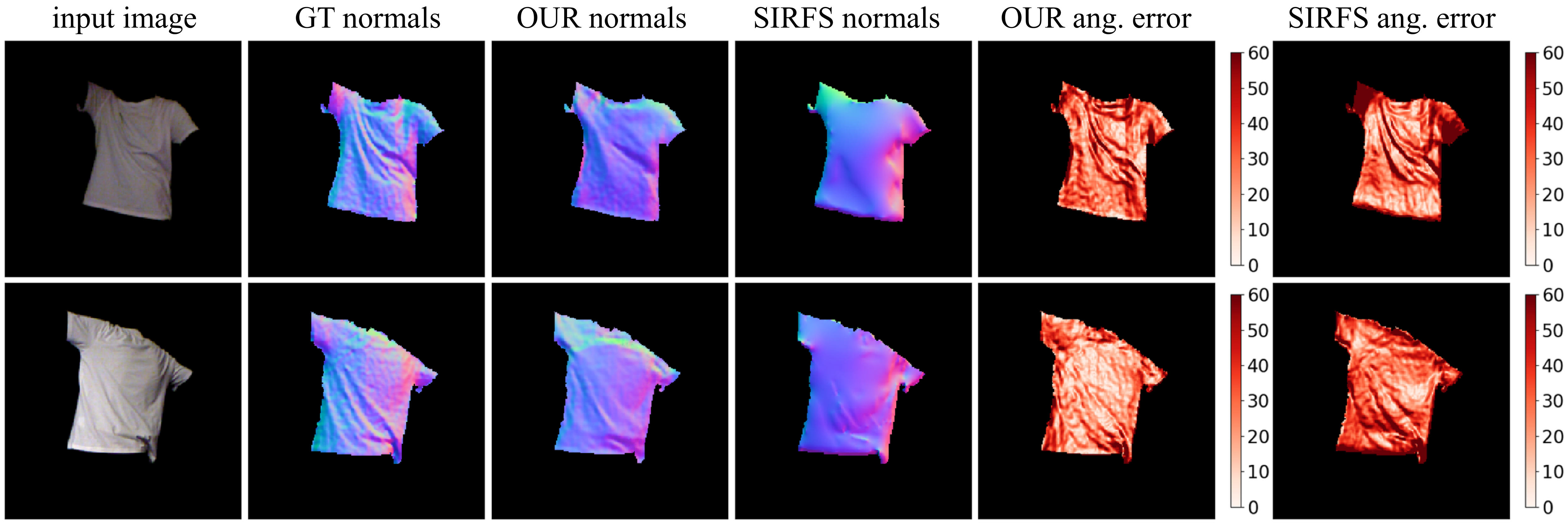}
	\caption{{\bf Qualitative comparison to SIRFS for the \texttt{cloth-tshirt} scenario.}  Even in this challenging scenario, our method is able to recover finer details than SIRFS.
	}
	\label{fig:normals_comp_L123d}
\end{figure*}

\begin{table*}[htbp]
	\centering
	\begingroup
	\setlength{\tabcolsep}{2pt}
	\renewcommand{\arraystretch}{0.8}
	\begin{tabular}{cccccccc}
		\multicolumn{1}{l}{\textbf{experiment}} & \textbf{method} & \multicolumn{1}{l}{\boldmath{}\textbf{mAE [$^{\circ}$]}\unboldmath{}} & \multicolumn{1}{l}{\boldmath{}\textbf{dAE [$^{\circ}$]}\unboldmath{}} & \multicolumn{1}{l}{\boldmath{}\textbf{$\mathbf{<10^{\circ}}$}\unboldmath{}} & \multicolumn{1}{l}{\boldmath{}\textbf{$\mathbf{<20^{\circ}}$}\unboldmath{}} & \multicolumn{1}{l}{\boldmath{}\textbf{$\mathbf{<30^{\circ}}$}\unboldmath{}} & \multicolumn{1}{l}{\boldmath{}\textbf{$m_{D}$ [mm]}\unboldmath{}} \\
		\midrule
		\multirow{2}[2]{*}{\textbf{cloth-cloth}} & SIRFS & $37.98\pm23.18$ & 33.52 & 7.25  & 24.93 & 43.96 & $31.55\pm10.93$ \\
		& OURS  & $\mathbf{17.37\pm12.51}$ & \textbf{14.44} & \textbf{30.6} & \textbf{68.85} & \textbf{87.29} & $\mathbf{17.53\pm5.50}$ \\
		\midrule
		\multirow{2}[2]{*}{\textbf{tshirt-tshirt}} & SIRFS & $30.17\pm20.26$ & 25.53 & 11.78 & 36.63 & 59.62 & $31.09\pm15.03$ \\
		& OURS  & $\mathbf{18.07\pm12.71}$ & \textbf{15.17} & \textbf{28.28} & \textbf{66.27} & \textbf{85.85} & \multicolumn{1}{l}{$\mathbf{17.18\pm18.58}$} \\
		\midrule
		\multirow{2}[2]{*}{\textbf{cloth-tshirt}} & SIRFS & $30.08\pm19.43$ & 25.93 & 10.49 & 35.03 & 59.15 & ${30.29\pm10.42}$ \\
		& OURS  & $\mathbf{25.74\pm15.81}$ & \textbf{22.81} & \textbf{13.45} & \textbf{41.98} & \textbf{67.7} & $\mathbf{26.26\pm7.72}$ \\
		\midrule
		\multirow{2}[2]{*}{\textbf{cloth-sweater}} & SIRFS & $33.25\pm21.60$ & 28.11 & 8.94  & 30.7  & 54.02 & $39.51\pm14.96$ \\
		& OURS  & $\mathbf{31.52\pm19.07}$ & \textbf{28.06} & \textbf{9.25} & \textbf{30.97} & \textbf{54.25} & $\mathbf{38.93\pm10.36}$ \\
		\midrule
		\multirow{2}[2]{*}{\textbf{cloth-hoody}} & SIRFS & $36.84\pm23.14$ & 32.11 & 7.79  & 26.2  & 46.11 & $43.51\pm13.79$ \\
		& OURS  & $\mathbf{32.54\pm21.15}$ & \textbf{28.02} & \textbf{9.88} & \textbf{31.78} & \textbf{54.05} & $\mathbf{43.22\pm24.81}$ \\
		\midrule
		\multirow{2}[2]{*}{\textbf{cloth-paper}} & SIRFS & $56.69\pm27.09$ & 59.53  & 1.71  & 7.06  & 15.73 & $49.35\pm18.51$ \\
		& OURS  & $\mathbf{35.53\pm22.16}$ & \textbf{31.13} & \textbf{8.42} & \textbf{27.54} & \textbf{47.84} & $\mathbf{24.16\pm7.15}$ \\
		\bottomrule
	\end{tabular}%
	\endgroup
	\vspace{0.2cm}
	\caption{{\bf Comparison with the method of~\cite{Barron15}.} Our approach outperforms SIRFS in all metrics, even in the challenging \texttt{cloth-tshirt}, \texttt{cloth-sweater}, \texttt{cloth-hoody} and \texttt{cloth-paper}  scenarios, with a particularly large gap for the first 5 metrics that evaluate the quality of the predicted normals. We report mean values averaged over test sets consisting of 100 samples each with standard deviations for the \mae{} and \md{} metrics.}
	\label{tab:comp_sirfs}%
	\vspace*{-0.4cm}
\end{table*}

The results of this comparison are summarized in Table~\ref{tab:comp_sirfs}, where we evaluate several normal-based error metrics and one depth-based metric, the \md{}. For the latter, we report the best results of SIRFS obtained either directly from the depth estimates, or by integrating the normal estimates. For our approach, we report the results based on the predicted normals, since we have found that they were slightly better than those obtained from our depth predictions. Note that we outperform SIRFS in all metrics. In the most challenging scenarios, \texttt{cloth-tshirt}, \texttt{cloth-sweater}, \texttt{cloth-hoody} and \texttt{cloth-paper}, our method still achieves lower errors, particularly in metrics evaluating normal quality. Figs.~\ref{fig:teaser} and~\ref{fig:normals_comp_L123d} show qualitative results for normal prediction on \tshirt{}. Our method clearly outperforms SIRFS when it comes to reconstructing the finer details of local creases. Furthermore, if we train our model on the combined dataset of \cloth{} and \tshirt{}, the generalization capability for objects of different categories drastically increases as is shown in Table \ref{tab:trained_on_cloth_tshirt}.

In Table~\ref{tab:comp_barron_time}, we compare the run-times of our approach with those of SIRFS. Note that our method performs \textit{orders of magnitude faster}. This is due to the fact that SIRFS relies on a costly optimization, whereas, in our case, all the heavy-lifting was done at training time and inference only requires a feed-forward pass through the network.

\begin{table}[t]
  \centering
	\resizebox{0.49\textwidth}{!}{
	\begingroup
	\setlength{\tabcolsep}{1pt}
	\renewcommand{\arraystretch}{1.1}
    \begin{tabular}{lcccccc}
    \textbf{Test set} & \boldmath{}\textbf{mAE [$^{\circ}$]}\unboldmath{} & \boldmath{}\textbf{dAE [$^{\circ}$]}\unboldmath{} & \boldmath{}\textbf{$\mathbf{<10^{\circ}}$}\unboldmath{} & \boldmath{}\textbf{$\mathbf{<20^{\circ}}$}\unboldmath{} & \boldmath{}\textbf{$\mathbf{<30^{\circ}}$}\unboldmath{} & \boldmath{}\textbf{$m_{D}$ [mm]}\unboldmath{} \\
    \midrule
    \textbf{sweater} & $25.75\pm16.72$ & 22.18 & 14.35 & 43.81 & 68.45 & $\mathbf{28.36\pm7.59}$ \\
    \textbf{hoody} & $24.66\pm17.36$ & 20.5  & 17.53 & 48.6  & 71.49 & $\mathbf{25.40\pm5.07}$ \\
    \bottomrule
    \end{tabular}%
	\endgroup
	}

\caption{{\bf Evaluation of our model trained on the combined \cloth{} and \tshirt{} dataset.} Exposing the model not only to the generic cloth but also to the T-Shirt worn by a person at training time helps the model learn a better shapes distribution which significantly improves the predictions when tested on different garment pieces (compare these results with the corresponding ones in Table~\ref{tab:comp_sirfs}).}
	\label{tab:trained_on_cloth_tshirt}%
		\vspace{-0.4cm}
\end{table}%

\begin{table}[t]
	\centering
	\begin{tabular}{ccccc}
		\textbf{Model} & SIRFS & OUR\_N & OUR\_D & OUR\_N+D \\
		\midrule
		\textbf{t [s]} & 113.653 & \textbf{0.01}  & \textbf{0.01}  & 0.016 \\
	\end{tabular}%
		\vspace{0.2cm}
\caption{{\bf Comparison of the run-times of our approach with those of SIRFS.} We report the average time needed to process one input image of size $224\times 224$~px.}
	\label{tab:comp_barron_time}
	\vspace*{-0.5cm}
\end{table}
\vspace*{-0.5cm}


\section{Conclusion} \label{sec:conclusion}
We have introduced a framework for reconstructing the 3D shape of a texture-less, deformable surface from a single image. To this end, we have followed a data-driven approach, thus essentially learning to perform Shape-from-Shading. Our experiments have demonstrated that, while meshes have proven effective to deal with well-textured deformable surfaces, they are much less well-suited than depth- and normal-based representations for texture-less ones in our setting. Furthermore, our comparison with a state-of-the-art SfS method has shown that our reconstructions were more accurate, particularly in terms of normal quality. This is the case even when training our model on one object and testing it on a different one. We expect that such a generalizability would further increase were we to use larger amounts of training data. Therefore, in the future, we will dedicate time to creating a larger-scale dataset of texture-less, deformable objects.
\paragraph{Acknowledgments}
This work was supported in part by a Swiss National Foundation for Research grant.

{\small
\bibliographystyle{ieee}
\bibliography{string,vision,learning}
}

\end{document}